\documentclass[conference]{IEEEtran}

     \PassOptionsToPackage{numbers, compress}{natbib}
\usepackage[utf8]{inputenc} %
\usepackage[T1]{fontenc}    %
\usepackage{url}            %
\usepackage{booktabs}       %
\usepackage{dcolumn}        %
\usepackage{amsfonts}       %
\usepackage{nicefrac}       %
\usepackage{microtype}      %
\usepackage{mathrsfs}
\usepackage{amsmath}
\usepackage{amssymb}
\usepackage{mathtools}
\usepackage{graphics}
\usepackage{float}
\usepackage{epstopdf}
\usepackage{graphicx}
\usepackage{bm}
\usepackage{amsthm}
\usepackage{tikz}
\usepackage[lined,boxed,commentsnumbered,linesnumbered, ruled]{algorithm2e}
\usepackage{comment}
\usepackage{balance}
\usepackage{cite}
\usepackage[capitalize]{cleveref}
\usepackage[size=small]{caption}
\usepackage[size=small,labelformat=simple]{subcaption}
\usepackage{ifthen}
\usepackage{csquotes}
\crefname{equation}{\unskip}{\unskip}
\usetikzlibrary{shapes, patterns,decorations.text, decorations.pathreplacing}

\newtheorem{proposition}{Proposition}

\newcommand*{\Scale}[2][4]{\scalebox{#1}{\ensuremath{#2}}}%

\newcommand{\fxp}[2]{\mathbb Q_{\langle #1, #2\rangle}}

\DeclareMathOperator*{\argmin}{arg\,min}
\DeclareMathOperator*{\supp}{supp}

\definecolor{darkgreen}{rgb}{0, 0.625, 0}

\title{Coding for Straggler Mitigation\\in Federated Learning}

\author{\IEEEauthorblockN{Siddhartha Kumar\IEEEauthorrefmark{1}, Reent Schlegel\IEEEauthorrefmark{1}, Eirik Rosnes\IEEEauthorrefmark{1}, and Alexandre Graell i Amat\IEEEauthorrefmark{2}\IEEEauthorrefmark{1}}
\IEEEauthorblockA{\IEEEauthorrefmark{1}Simula UiB, Bergen, Norway
}\IEEEauthorblockA{\IEEEauthorrefmark{2}Department of Electrical Engineering, Chalmers University of Technology, Gothenburg, Sweden
}

\thanks{This work was financially supported by the Swedish Research Council under grant 2020-03687.}}

\IEEEoverridecommandlockouts 
\begin{document}

\maketitle

\begin{abstract}
    We present a novel  coded federated learning (FL) scheme for linear regression that mitigates the effect of straggling devices while retaining the privacy level of conventional FL.  The proposed scheme combines  one-time padding to preserve privacy and gradient codes to yield resiliency against stragglers and consists of two phases.
    In the first phase, the devices share a one-time padded version of their local data with a subset of other devices. In the second phase, the devices and the central server collaboratively and iteratively train a global linear model using gradient codes on the one-time padded local data. To apply one-time padding to real data, our scheme exploits a fixed-point  arithmetic representation of the  data.
    Unlike the coded FL scheme recently introduced by Prakash \emph{et al.}, the proposed scheme maintains the same level of privacy as conventional FL while achieving a similar training time. 
    Compared to conventional FL, we show that the proposed scheme  achieves a training speed-up factor of $6.6$ and $9.2$ on the MNIST and Fashion-MNIST datasets for an accuracy of $95\%$ and $85\%$, respectively.
\end{abstract}

\section{Introduction}
Federated learning (FL) \cite{McM17,Kon16,Tian20} is a distributed learning paradigm that
trains an algorithm across multiple devices without exchanging  the training data directly, thus limiting the privacy leakage and reducing the communication load. 
In many applications of FL, such as in the Internet of Things (IoT), due to the heterogeneous nature of the training devices and instability of the communication links, the  training latency can be severely impaired by \emph{straggling devices}, i.e., devices that do not provide timely updates. 
Various FL algorithms have been proposed in the literature to tackle stragglers. The most popular is federated averaging \cite{McM17}, which mitigates the effect of stragglers by dropping the slowest devices at the cost of reduced accuracy. When data is non-identically distributed across devices, which is typically the case in practice, the loss in accuracy may be significant\textemdash in this case, dropping stragglers makes the algorithm suffer from the client drift phenomenon, i.e., the learning converges to the optimum of one of the local models \cite{Cha20,Mit21}. 
Straggler mitigating schemes for scenarios for which the data is identically distributed across devices were presented in \cite{Dut18, Rei20}, while the authors of \cite{Mit21,Li20,Wan20,Wu21} introduced asynchronous schemes to deal with scenarios for which the data is non-identically distributed across devices. The key idea here is to make use of  stale information (e.g., gradients) from the stragglers rather than discarding them at the central server. Generally, schemes of such nature do not converge to the global optimum. In particular, the authors of \cite{Li20} presented a scheme that controls the client drift, but with a nonlinear convergence rate to the global optimum \cite{Mit21}.

Mitigating the impact of stragglers has also been addressed in the neighboring area of distributed computing in data centers for matrix-vector and matrix-matrix multiplication \cite{Li16,Yu2017,Lee18,Sev19,amir2019}, distributed gradient descent \cite{Tan17}, and distributed optimization \cite{Kar17}, as well as in the context of edge computing \cite{Sch22,Zha19,Fri21} and FL \cite{Pra21}. The key idea  is to  introduce redundant computations by means of an erasure correcting code\textemdash thereby increasing the computational load at each server\textemdash so that the result of a computation task can be obtained from the subtasks completed by a subset of the servers. In FL, the fact that the raw data is distributed across devices beforehand precludes from introducing redundant computations in the same manner as in distributed computing.
The main idea in \cite{Pra21} is that devices generate parity data, which is shared with the central server to facilitate the training and provide resilience against straggling devices. The sharing of the parity (coded) data with the central server, however, leaks information of the raw data to the central server, i.e., the coded FL scheme in \cite{Pra21} yields a lower level of privacy than conventional FL.

In this paper, we propose a novel privacy-preserving coded FL scheme for linear regression that mitigates the effect of straggling devices and converges to the global optimum. Hence, the proposed scheme yields no  penalty on the accuracy even for highly non-identically distributed data across devices. Furthermore, unlike the scheme in \cite{Pra21}, it retains the privacy level of conventional FL against the central server and honest-but-curious devices. 
The scheme consists of two phases: in the first phase, the devices share a one-time padded version of their local data with a subset of other devices. The sharing of one-time padded data does not reveal any information about the data to other devices but  enables the use of erasure correcting codes in the second phase. Particularly, in this phase, each device uses a gradient code \cite{Tan17} to generate a partial gradient on the local data and the padded data received from other devices. The partial gradient is then shared with the central server, which aggregates the received partial gradients\textemdash after removing the random keys\textemdash and sends an updated global model to the devices. 
We show that, for a realistic IoT environment, the proposed coded FL scheme using kernel embedding (for linearization) achieves a speed-up factor of $6.6$ and $9.2$ compared to conventional FL when training on the MNIST \cite{Cun10} and Fashion-MNIST \cite{Xia17}  datasets for an accuracy of $95\%$ and $85\%$, respectively. 

\emph{Notation.} We use uppercase and lowercase bold letters for matrices and vectors, respectively, italics for sets, and uppercase sans-serif letters for random variables, e.g., \(\bm X\), \(\bm x\),  \(\mathcal{X}\), and \(\mathsf{X}\)  represent a matrix, a vector, a set, and a random variable, respectively. An exception to this rule is $\bm \epsilon$, which will denote a matrix. Vectors are represented as row vectors throughout the paper. For natural numbers  \(c\) and \(d\), \(\bm 1_{c\times d}\) denotes an all-one matrix of size \(c\times d\). The transpose of a matrix \(\bm X\) is denoted as \(\bm X^\top\). The support of a vector $\bm x$ is denoted by $\supp(\bm x)$, while the gradient of a function $f(\bm X)$ with respect to $\bm X$ is denoted by $\nabla_{\bm X}f(\bm X)$. Furthermore, we represent the Euclidean norm of a vector $\bm x$ by $\Vert \bm x \Vert$, while the Frobenius norm of a matrix \(\bm X\) is denoted by   \(\Vert\bm X\Vert_\text{F}\).  Given integers \(a, b\in\mathbb{Z}\), \(a<b\), we define \([a,b]\triangleq\{a,\ldots,b\}\), where \(\mathbb{Z}\) is the set of integers, and \([a]\triangleq\{1,\ldots, a\}\) for a positive integer $a$. Additionally, %
for a real number $e$, $\lfloor e \rfloor$ is the largest integer less than or equal to $e$.  The expectation of a random variable $\mathsf{\Lambda}$ is denoted by $\mathbb{E}[{\mathsf\Lambda}]$, and we write $\mathsf{\Lambda} \sim  \text{geo}(1-p)$ to denote that $\mathsf{\Lambda}$ follows a geometric distribution with failure probability \(p\).

\section{Preliminaries}
\subsection{Fixed-Point Numbers} \label{sec:Fixed-Point Numbers}
Fixed-point numbers are rational numbers that can be split into an integer part and a fractional part. Let \(s\cdot\nolinebreak(\delta_{k-f-2}\dots\delta_0\,.\,\delta_{-1}\dots\delta_{-f})\) be the binary  representation of a fixed-point number $\tilde x$, of value \(\tilde x=s\cdot \sum_{i=-f}^{k-f-2}\delta_i2^i\), where  $s=\mathrm{sign}(\tilde x)$ is the sign of $\tilde x$,  $k-f$ is the length of the integer part (including the sign), and  $f$  the length of the fractional part. Also, let  \(\bar x=s\cdot\sum_{i=-f}^{k-f-2}\delta_i2^{i+f}\in\mathbb{Z}\). Then, \(\tilde x=\bar x 2^{-f}\), i.e.,  fixed-point numbers can be seen as integers  scaled by a factor \(2^{-f}\). Let \(\mathbb Z_{\langle k \rangle}=[-2^{k-1}, 2^{k-1}-1]\). We define the set  \(\fxp{k}{f}\triangleq\{\tilde x =\bar x 2^{-f}, \bar x \in \mathbb Z_{\langle k \rangle}\}\) of all fixed-point numbers with range \(2^{k-f}\) and resolution \(2^{-f}\).

\subsection{Cyclic Gradient Codes}
Gradient codes \cite{Tan17} are a class of codes designed to mitigate the effect of stragglers in distributed gradient descent in data centers. Consider a piece of data  partitioned into \(D\) partitions, which are distributed among \(D\) servers, each storing \(\alpha\leq D\) partitions. An \((\alpha, D)\) fixed-point gradient code is characterized by the matrices \(\bm A\in\fxp{k}{f}^{S\times D}\) and \(\bm B\in\fxp{k}{f}^{D\times D}\) of size \(S\times D\) and \(D\times D\), respectively, where \(S\) denotes the number of straggling patterns that the code can deal with. The \(i\)-th row of \(\bm B\), \(i\in[D]\), is associated with the \(i\)-th server\textemdash the support of the row corresponds to the partitions of the data assigned to that  server. Furthermore, we assume that the supports of the rows of \(\bm B\), each of size $\alpha$, follow a cyclic pattern. 
We will refer to such cyclic gradient codes simply as gradient codes. Now, let \(\bm g_i\in\fxp{k}{f}^d\) denote the gradient of dimension $d$ of the \(i\)-th partition. The encoding of the gradients at each server is given by $\bm B\left(\bm g_1^\top,\ldots, \bm g_D^\top\right)^\top$, where the $i$-th row corresponds to server $i$. Each server then sends the encoding of the local gradients to a master server, whose aim is to linearly combine any \(D-\alpha+1\) of them to obtain  \(\sum_i \bm g_i\),  thus mitigating the impact of stragglers. We refer to this operation as the \emph{decoding} operation, which  is determined by \(\bm A\). Particularly, if the master server receives gradients from a subset of servers $\mathcal{A}=\{a_1,\ldots,a_{|\mathcal A|}\}$, it applies the linear combination of these gradients with coefficient for server $a_i$ given by the $a_i$-th element of the row of $\bm A$ with support $\mathcal{A}$.
Moreover, it is required that 
\begin{align}
    \label{Eq: GradientCodeCondition}
    \bm A\bm B=\bm 1_{S\times D}\,.
\end{align}

We refer the interested reader to \cite[Alg.~1]{Tan17} and \cite[Alg.~2]{Tan17} for the construction of \(\bm A\) and \(\bm B\), respectively.

\section{System Model}

We consider an FL scenario in which  \(D\) devices collaborate to train a machine learning model with the help of a central server. Device  \(i\in[D]\) has local data \( \mathcal{D}_i=\{(\bm x_j^{(i)}, \bm y_j^{(i)})\mid  j\in[n_i]\} \) consisting of \(n_i\) training points. We denote by \(m\)  the total number of data points across all devices, i.e., \(m=\sum_i n_i\).
The  scheme proposed in Section~\ref{Sec: CodedFederatedBatchGradient} is based on one-time padding, which cannot be applied over the reals. To circumvent this shortcoming, our scheme works on the fixed-point representation of the data. Hereafter, we assume that \(\bm x_j^{(i)}\in\fxp{k}{f}^d\) and \(\bm y_j^{(i)}\in\fxp{k}{f}^c\) are the  fixed-point representations of the corresponding real-valued vectors. Note that practical systems operate in fixed-point, hence the proposed scheme does not incur in a limiting assumption.

 We represent the data in matrix form  as
\begin{align*}
    \bm X^{(i)} &= \left(\begin{matrix}
        \bm x_1^{(i)\top},\ldots,\bm x_{n_i}^{(i)\top}
    \end{matrix}\right)^\top\,,\\
    \bm Y^{(i)}&= \left(\begin{matrix}
        \bm y_1^{(i)\top},\ldots,\bm y_{n_i}^{(i)\top}
    \end{matrix}
    \right)^\top\,,
\end{align*}
where \(\bm X^{(i)}\) is of size \(n_i\times d\) and \(\bm Y^{(i)}\)  of size \(n_i\times c\). The devices and the central server collaboratively try to infer a linear global model \(\bm \Theta\in\fxp{k}{f}^{d\times c}\) as
$\bm y=\bm x\bm \Theta$, 
where  \(\bm x\) is a feature vector and \(\bm y\) its corresponding label, using  federated gradient descent.

\subsection{Federated Gradient Descent}
For convenience, we collect the whole data (consisting of \(m\) data points) in matrices $\bm X$ and $\bm Y$  as
\begin{align*}
    \bm X=\left(\begin{matrix}
        \bm x_1\\
        \vdots\\
        \bm x_m
    \end{matrix}\right)=\left(\begin{matrix}
        \bm X^{(1)}\\
        \vdots\\
        \bm X^{(D)}
    \end{matrix}\right)
    \;\,\text{and}\;\,
    \bm Y=\left(\begin{matrix}
        \bm y_1\\
        \vdots\\
        \bm y_{m}
    \end{matrix}\right)
    =\left(\begin{matrix}
        \bm Y^{(1)}\\
        \vdots\\
        \bm Y^{(D)}
    \end{matrix}\right)\,,
\end{align*}
where \(\bm X\) is of size \(m\times d\) and \(\bm Y\)  of size  \(m\times c\).
Inferring the linear model $\bm \Theta$ can be formalized as the minimization problem 
\begin{align}
\label{Eq: OptimisationProblem}
    \argmin_{\bm \Theta}\; f(\bm \Theta)\triangleq\frac{1}{2m}\sum_{l=1}^m \left\Vert \bm x_l\bm \Theta-\bm y_l \right\Vert^2+\frac{\lambda}{2}\left\Vert\bm \Theta\right\Vert^2_\text{F}\,,
\end{align}
where $f(\bm \Theta)$ is the \emph{global} loss function and \(\lambda\)  the regularization parameter.

Let \(f_i(\bm \Theta)\) denote the \emph{local} loss function corresponding to the data at device \(i\), i.e.,
\(f_i(\bm \Theta)=\frac{1}{2n_i}\sum_{j=1}^{n_i}\Vert\bm x_j^{(i)}\bm \Theta- \bm y_j^{(i)}\Vert^2.\)
Then, \(f(\bm \Theta)\) in \cref{Eq: OptimisationProblem} can be expressed as
\(
    f(\bm \Theta)=\sum_{i=1}^{D}\frac{n_i}{m}f_i(\bm \Theta) + \frac{\lambda}{2}\left\Vert\bm \Theta\right\Vert^2_\text{F}.
\)

Federated gradient descent proceeds iteratively to train the model  \(\bm \Theta\). At each epoch, the devices compute the gradients of the respective loss functions and send them to the central server, which  aggregates the received gradients to update the model. More precisely, during the \(e\)-th epoch,  device $i$ computes the gradient
\begin{align}
    \label{Eq: DeviceComputation}
    \Scale[1.0]{\bm G_i^{(e)}= n_i\nabla_{\bm \Theta} f_i(\bm \Theta^{(e)})=%
    {\bm X^{(i)\top}}\bm X^{(i)}\bm \Theta^{(e)}-{\bm X^{(i)\top}} \bm Y^{(i)}}\,,
\end{align}
where $\bm \Theta^{(e)}$ denotes the current model estimate. 
Upon reception of the gradients, the central server aggregates them to update the model according to
\begin{align}
    \label{Eq: ServerAggregation}
    \nabla_{\bm \Theta} f(\bm \Theta^{(e)})&=\sum_{i=1}^D %
    \frac{1}{m}\bm G_i^{(e)}+\lambda\bm \Theta^{(e)}\,,\\
    \label{Eq: ServerUpdate}
    \bm \Theta^{(e+1)}&=\bm \Theta^{(e)}-\mu\nabla_{\bm \Theta} f(\bm \Theta^{(e)})\,,
\end{align}
where \(\mu\) is the learning rate. The updated model  \(\bm \Theta^{(e+1)}\) is then sent back to the devices, and \cref{Eq: DeviceComputation,Eq: ServerAggregation,Eq: ServerUpdate} are iterated $E$ times until convergence, i.e., until \(\bm \Theta^{(E+1)}\approx\bm \Theta^{(E)}\).

\subsection{Computation and Communication Latency}

Let \(\mathsf{T}^{\mathsf{comp}}_i\) be the  time required to compute \(\rho_i\) multiply and accumulate (MAC) operations by device $i$. Similar to \cite{Zha19}, we model \(\mathsf{T}^{\mathsf{comp}}_i\) as a shifted exponential random variable,
\begin{align*}
\mathsf{T}^{\mathsf{comp}}_i= \frac{\rho_i}{\tau_i}+\mathsf{\Lambda_i}\,,
\end{align*}
where  \(\{\mathsf{\Lambda}_i\}\) are independent  exponential random variables with \(\mathbb{E}[\mathsf{\Lambda}_i]=1/\eta_i\) representing the random setup times required by the devices, and \(\tau_i\) is the number of MAC operations per second performed by device $i$.

We assume that communication  between the central server and the devices may fail. To enable communication, the devices and the central server repetitively transmit during the uplink and  downlink phases until the first successful transmission occurs. Let \(\mathsf{N}_i^{\mathsf{u}}\sim\mathrm{geo}\,(1-p_i)\) and \(\mathsf{N}_i^{\mathsf{d}}\sim\mathrm{geo}\,(1-p_i)\) denote the number of transmissions needed for successful communication in the uplink and downlink, respectively, where \(p_i\) denotes the failure probability of a single transmission between the central server and device $i$. Also, let \(\gamma^{\mathsf{u}}\) and \(\gamma^{\mathsf{d}}\) be the transmission rates between the central server and the devices in the uplink and downlink, respectively. Then, the time required to successfully communicate \(b\) bits during uplink and  downlink, denoted by $\mathsf{T}^{\mathsf{u}}_i$ and $\mathsf{T}^{\mathsf{d}}_i$, respectively, is 
\[\mathsf{T}^{\mathsf{u}}_i=\frac{\mathsf{N}^{\mathsf{u}}_i}{\gamma^{\mathsf{u}}}b \quad \text{and} \quad \mathsf{T}^{\mathsf{d}}_i=\frac{\mathsf{N}^{\mathsf{d}}_i}{\gamma^{\mathsf{d}}}b\,.\]

In our model, all communication between any two devices
happens over a secured link and is relayed through the central server, i.e., any two devices share an encrypted communication link and the central server learns nothing about the exchanged messages.

\section{Low-Latency Federated Gradient Descent}
\label{Sec: CodedFederatedBatchGradient}

The proposed scheme builds on one-time padding and gradient codes. Note, however, that one-time padding cannot be applied to data over the reals. To bypass this problem, we consider a fixed-point representation of the data and apply fixed-point arithmetic operations. In the following, we first discuss how to preserve privacy in performing operations using fixed-point arithmetic and then present  the proposed scheme.

\subsection{Privacy-Preserving Operations on Fixed-Point Numbers }

The authors of \cite{Cat10} were the first to address the problem of performing secure computations (in the context of multiparty computation) using fixed-point numbers. The idea is to map fixed-point numbers to finite-field elements, and then perform secure operations (addition, multiplication, and division) of two secretly-shared numbers over the finite field. In this paper, we use a similar  approach as the one in \cite{Cat10} but define a different mapping and a simplified multiplication operation, leveraging the fact that we only need to multiply a secretly-shared number with a public number, as discussed in the next subsection. The resulting protocol is more efficient than the one in \cite{Cat10}.

Consider the  fixed-point datatype $\fxp{k}{f}$ (see \cref{sec:Fixed-Point Numbers}). Secure addition on $\fxp{k}{f}$ can be performed via simple integer addition with an additional modulo operation. Let $(\cdot)_{\mathbb Z_{\langle k \rangle}}$ be the map from the integers onto the set $\mathbb Z_{\langle k \rangle}$ given by the modulo operation. Furthermore, let $\tilde{a},\tilde{b}\in\fxp{k}{f}$, with $\tilde{a} = \bar{a}2^{-f}$ and $\tilde{b} = \bar{b}2^{-f}$. For $\tilde{c} = \tilde{a} + \tilde{b}$, with $\tilde{c} = \bar{c}2^{-f}$, we have $\bar{c} = (\bar{a} + \bar{b})_{\mathbb Z_{\langle k \rangle}}$.

Multiplication on $\fxp{k}{f}$ is performed via integer multiplication with scaling over the reals in order to retain the precision of the datatype and an additional modulo operation. For $\tilde{d} = \tilde{a} \cdot \tilde{b}$, with $\tilde{d} = \bar{d}2^{-f}$, we have $\bar{d} = (\lfloor\bar{a}\cdot\bar{b}\cdot 2^{-f}\rfloor)_{\mathbb Z_{\langle k \rangle}}$.

\begin{proposition}[Perfect privacy]
    \label{Th: Perfect Privacy}
    Consider a secret \(\tilde{x}\in\fxp{k}{f}\) and a key \(\tilde r\in\fxp{k}{f}\) that is picked uniformly at random. Then, \(\tilde{x}+\tilde{r}\) is uniformly distributed in \(\fxp{k}{f}\), i.e., \(\tilde{x}+\tilde{r}\) does not reveal any information about \(\tilde{x}\).
\end{proposition}

\cref{Th: Perfect Privacy} is an application of a one-time pad, which was proven secure by Shannon in \cite{Shannon49}.

\begin{proposition}[Retrieval]
    Consider a public fixed-point number \(\tilde{c}\in\fxp{k}{f}\), a secret \(\tilde{x}\in\fxp{k}{f}\), and a key \(\tilde{r}\in\fxp{k}{f}\) that is picked uniformly at random. Suppose we have the weighted sum \(\tilde{c}(\tilde{x}+\tilde{r})\) and the key. Then, we can retrieve \(\tilde{c}\tilde{x}=\tilde{c}(\tilde{x}+\tilde{r})-\tilde{c}\tilde{r}+O(2^{-f})\).
\end{proposition}

The above proposition tells us that, given \(\tilde{c}\), \(\tilde{c}(\tilde{x}+\tilde{r})\), and \(\tilde{r}\), it is possible to obtain an approximation of \(\tilde{c}\tilde{x}\). Moreover, if we choose a sufficiently large  \(f\),  then we can retrieve \(\tilde{c}\tilde{x}\) with negligible error.

\subsection{Data Sharing Scheme}

We are now ready to introduce the proposed privacy-preserving scheme. It consists of two phases: in the first phase, discussed in this subsection, we secretly share data between devices, which enables the  use of gradient codes in the second phase to perform privacy-preserving coded federated gradient descent while conferring straggler mitigation. %

The central server first generates two sets of keys,  \(\mathcal{K}_1=\{\bm \Delta_{1},\ldots,\bm \Delta_{D}\}\) and \(\mathcal{K}_2=\{\bm\Xi_{1},\ldots,\bm\Xi_{D}\}\), where  \(\bm \Delta_{i}\) and \(\bm \Xi_{i}\) are sent to device \(i\),\footnote{Note that we consider the communication cost of transmitting keys to be negligible since, in practice, it is enough to send a (much smaller) pseudorandom number generator seed instead of the random numbers.} \(\bm \Delta_i\) is a matrix of size  \(d\times c\), and  \(\bm \Xi_{i}\) is a symmetric matrix of size $d \times d$. Using its keys and its data \(\mathcal{D}_i\), device \(i\) computes
\begin{align}
    \bm \Psi_{i}&= \bm G_i^{(1)}+\bm \Delta_{i}\,, %
    \label{Eq: PaddedData2} \\
    \bm\Phi_{i}&=%
    {\bm X^{(i)\top}} \bm X^{(i)} + \bm \Xi_i\,,
    \label{eq:phi}
\end{align}
where $\bm G_i^{(1)}$ is the gradient of device $i$ in the first epoch (see~\cref{Eq: DeviceComputation}).

The above matrices are one-time padded versions of the gradient and transformed data. Sharing \(\bm \Psi_i\) and \(\bm \Phi_i\) does not leak any information about the data of device \(i\), but it is critical nevertheless, as it introduces redundancy of the data across devices, which enables the use of gradient codes in the second phase.
In the following, we describe the sharing process.

Let \(\alpha\leq D\) be the number of local datasets to be stored at each device (including its own), and $\bm B$ the encoding matrix of an $(\alpha,D)$ gradient code in fixed-point representation with entries $b_{ij}$. Each device $i$ has to send $\bm \Psi_{i}$ and $\bm \Phi_{i}$ to $\alpha - 1$ other devices given by the support of the $i$-th row of $\bm B$. We  denote the support of the $i$-th row of $\bm B$ as $\{\omega_{1i},\ldots,\omega_{\alpha i}\}$. Note that $i$ is always in the support of row $i$ and the devices do not have to send data to themselves. Hence, each device shares its padded data and gradient with only $\alpha -1$ other devices. We assume that the devices are equipped with full-duplex technology and have orthogonal channels to the central server, i.e., the devices can share their data simultaneously.

Once the padded gradient and data have been shared, device $i$ computes
\begin{align}
    \bm C_{i} &=\left(
        b_{i,\omega_{1i}}, \ldots, b_{i, \omega_{\alpha i}}
    \right)\Scale[1.0]{\left(
        \bm \Psi_{\omega_{1i}}^\top\,,
        \ldots,
        \bm \Psi_{\omega_{\alpha i}}^\top
    \right)^\top},
    \label{eq:C}\\
    \bar{\bm C}_{i} &=\left(
        b_{i,\omega_{1i}}, \ldots, b_{i, \omega_{\alpha i}}
    \right)\Scale[0.96]{\left(
        \bm \Phi_{\omega_{1i}}^\top\,,
        \ldots, 
        \bm \Phi_{\omega_{\alpha i}}^\top
    \right)^\top}\,,
    \label{eq:Cbar}
\end{align}
which completes the sharing phase. Equation \cref{eq:C} corresponds to the encoding at device $i$ via a gradient code of the padded gradients at epoch $1$. Similarly,  \cref{eq:Cbar} corresponds to the encoding at device $i$ of the available padded data.

\subsection{Coded Gradient Descent}

After the transmission phase, the central server and the devices iteratively train a global model using gradient descent. Consider the \(e\)-th epoch and let
\begin{equation}
    \bm \Theta^{(e)}=\bm \Theta^{(1)}+\bm\epsilon^{(e)}
    \label{eq:theta}
\end{equation}
be the model parameter at the  $e$-th epoch, where   \(\bm\epsilon^{(e)}\) is the update matrix and \(\bm \Theta^{(1)}\)  the initial model estimate. Instead of sending \(\bm \Theta^{(e)}\) to the devices, as is standard for gradient descent, in the proposed coded gradient descent, the central server sends the update matrix \(\bm\epsilon^{(e)}\).%

Upon reception of \(\bm\epsilon^{(e)}\), the devices compute the gradients $\tilde{\bm G}_i^{(e)}$ on the encoded padded data. Particularly,  in the \(e\)-th epoch,  device \(i\) computes the gradient
\begin{align*}
\tilde{\bm G}_i^{(e)}  &= \bm C_{i}+ \bar{\bm C}_{i} \bm \epsilon^{(e)}\\
&\overset{(a)}{=} \sum_{j=1}^\alpha b_{i, \omega_{ji}}\bigg(\bm G_{\omega_{ji}}^{(1)} + \bm \Delta_{\omega_{ji}}\bigg)\\
&\quad\quad+\sum_{j=1}^\alpha b_{i, \omega_{ji}} \bigg({\bm X^{(\omega_{ji})\top}} \bm X^{(\omega_{ji})} + \bm \Xi_{\omega_{ji}} \bigg)\bm \epsilon^{(e)}\\
&\overset{(b)}{=} \sum_{j=1}^\alpha b_{i, \omega_{ji}}\bigg(\bm G_{\omega_{ji}}^{(1)} + {\bm X^{(\omega_{ji})\top}} \bm X^{(\omega_{ji})}\bm \epsilon^{(e)}\bigg)\\
&\quad\quad+ \sum_{j=1}^\alpha b_{i, \omega_{ji}} \bigg( \bm \Delta_{\omega_{ji}} + \bm \Xi_{\omega_{ji}}\bm \epsilon^{(e)}\bigg)\\
&\overset{(c)}{=} \sum_{j=1}^\alpha b_{i, \omega_{ji}} \bigg(\bm G_{\omega_{ji}}^{(e)}+\bm \Xi_{\omega_{ji}}\bm\epsilon^{(e)}+\bm \Delta_{\omega_{ji}}\bigg)\,,
\end{align*}
where $(a)$ follows from \cref{eq:C} and \cref{eq:Cbar} together with \cref{Eq: PaddedData2} and \cref{eq:phi}, $(b)$ is a reordering, and $(c)$ follows from \cref{Eq: DeviceComputation} and \cref{eq:theta}. Device $i$ then sends $\tilde{\bm G}_i^{(e)}$ to the central server, which updates the global model as explained next.
The central server waits for the first \(D-\alpha+1\) gradients it receives, subtracts the keys (as it knows \(\bm B\) and the keys \(\mathcal{K}_1\) and \(\mathcal{K}_2\)), and performs a decoding operation based on matrix \(\bm A\), where $\bm A$ is the decoding matrix for the gradient code given by $\bm B$. Let \(\mathcal{A}\subset[D]\), \(|\mathcal{A}|=D-\alpha+1\), be the set of indices of the $D-\alpha+1$ fastest devices to finish the computation of $\tilde{\bm G}_i^{(e)}$. After removing the keys from $\tilde{\bm G}_i^{(e)}$, \(\forall i\in\mathcal{A}\), the central server obtains
\(\bm P_i^{(e)}=\sum_{j=1}^\alpha b_{i, \omega_{ji}} \bm G_{\omega_{ji}}^{(e)}.\)
Next, it decodes according to \(\bm A\) as follows. Let \(\bm a_s=(a_{s,1}, a_{s,2}, \ldots, a_{s,D})\) be the \(s\)-th row of \(\bm A\) such that \(\supp(\bm a_s)=\mathcal{A}\). Then,
\begin{equation}
    \label{Eq: CodedFL_UpdateRule}
    \sum_{i\in\mathcal{A}}a_{s,i}\bm P_i^{(e)}\overset{(a)}{=}\sum_{i=1}^D \bm G_i^{(e)}
    \overset{(b)}{=}m\big(\nabla_{\bm \Theta}f(\bm \Theta^{(e)}) - \lambda\bm\Theta^{(e)}\big)\,,
\end{equation}
where $(a)$ follows from the property of gradient codes in \cref{Eq: GradientCodeCondition} and $(b)$ follows from \cref{Eq: ServerAggregation}. Lastly, \(\bm \Theta^{(e+1)}\) is obtained according to \cref{Eq: ServerUpdate} for the next epoch. Note that for the central server to obtain the correct global model update, the devices can perform only one epoch of local training between two successive global updates. This restriction means that our scheme can only be applied to federated gradient descent and not to federated averaging, where devices perform multiple local model updates before the central server updates the global model.

\begin{proposition}
\label{prop:Prop}
    The proposed \((\alpha,D)\) coded FL scheme is resilient to \(\alpha - 1\) stragglers, and achieves the global optimum, i.e., the optimal model obtained through gradient descent for  linear regression.
\end{proposition}
\begin{IEEEproof}
    From \cref{Eq: CodedFL_UpdateRule}, we see that during each epoch, \(e\), the central server obtains
    \begin{align*}
    \nabla_{\bm \Theta}f(\bm\Theta^{(e)}) &= \frac{1}{m}\sum_{i=1}^D \bm G_i^{(e)} + \lambda\bm \Theta^{(e)} \\
    &=\frac{1}{m}\bm X^\top(\bm X\bm\Theta^{(e)} - \bm Y) + \lambda\bm \Theta^{(e)}\,,
    \end{align*}
    using the coded data obtained from the \(D-\alpha+1\) fastest devices. It further obtains an updated linear model using \cref{Eq: ServerUpdate}, which is exactly the update rule for gradient descent.
\end{IEEEproof}

\section{Numerical Results}

We simulate a wireless setting with \(D=25\) devices and a central server which want to perform FL on the MNIST \cite{Cun10} and Fashion-MNIST  \cite{Xia17}  datasets.
To simulate non-identically distributed data, we sort the training data corresponding to the labels and then we divide the training data into $D=25$ equal parts, one for each device. Each device  pre-processes its assigned data using kernel embedding as done by the radial basis function sampler of Python's sklearn package (with $5$ as kernel parameter and $2000$ features) to obtain the (random) features \(\bm X^{(i)}\) and then stores \({\bm X^{(i)\top}}\bm X^{(i)}\). We assume that the pre-processing step is performed offline. For conventional FL, the devices use 32-bit floating point arithmetic, whereas in the proposed coded FL scheme, the devices work on fixed-point numbers with \(k = 48\) bits out of which \(f=24\) bits are for the fractional part. Furthermore, for conventional FL the data at the devices is divided into five smaller batches and we perform mini-batch learning to speed up the process by reducing the epoch times. The mini-batch size is chosen as a compromise between the two corner cases: 
a mini-batch size of $1$ is difficult to parallelize, whereas a large mini-batch size may exceed the devices' limited parallelization capabilities leading to an increased computational latency at each epoch. We select a mini-batch size of $480$  as a middle ground, which allows to utilize the parallelization capabilities of the considered chips while keeping the computational load at each epoch reasonable.

We consider devices with heterogeneous computation capabilities, which we model by varying the MAC rates \(\tau_i\). In particular, we have four classes of devices: $10$ devices have a MAC rate of $25 \cdot 10^6$ MAC/s, $5$ devices have $5 \cdot 10^6$, $5$ have $2.5 \cdot 10^6$, and the last $5$ have $1.25 \cdot 10^6$, whereas the central server has a MAC rate of $8.24 \cdot 10^{12}$ MAC/s. We chose these MAC rates based on the performance that can be expected by using devices with chips from Texas Instruments of the TI MSP430 family \cite{TIchip}.
We sample the setup times $\mathsf{\Lambda_i}$ at each epoch and assume that they have an expected value of $50\%$ of the deterministic computation time, i.e., device $i$ performing $\rho_i$ MAC operations at each epoch yields $\eta_i = \frac{2\tau_i}{\rho_i}$.
The communication between the central server and the devices is based on the LTE Cat 1 standard for IoT applications and the corresponding rates are \(\gamma^{\mathsf{d}}=10\) Mbit/s and \(\gamma^{\mathsf{u}}=5\) Mbit/s. The probability of transmission failure between the central server and the devices is constant across devices, with \(p_i=0.1\), $\forall i \in [D]$, and we assume a header overhead of \(10\%\) for each packet. Lastly, we use regularization parameter \(\lambda=9\times10^{-6}\) and initial learning rate \(\mu=6\). The learning rate $\mu$ is updated as \(\mu\leftarrow0.8\mu\) at epochs $200$ and $350$.

\begin{figure}
    \centering
    \begin{subfigure}[b]{0.45\textwidth}
        \centering
        \includegraphics[width=1\columnwidth]{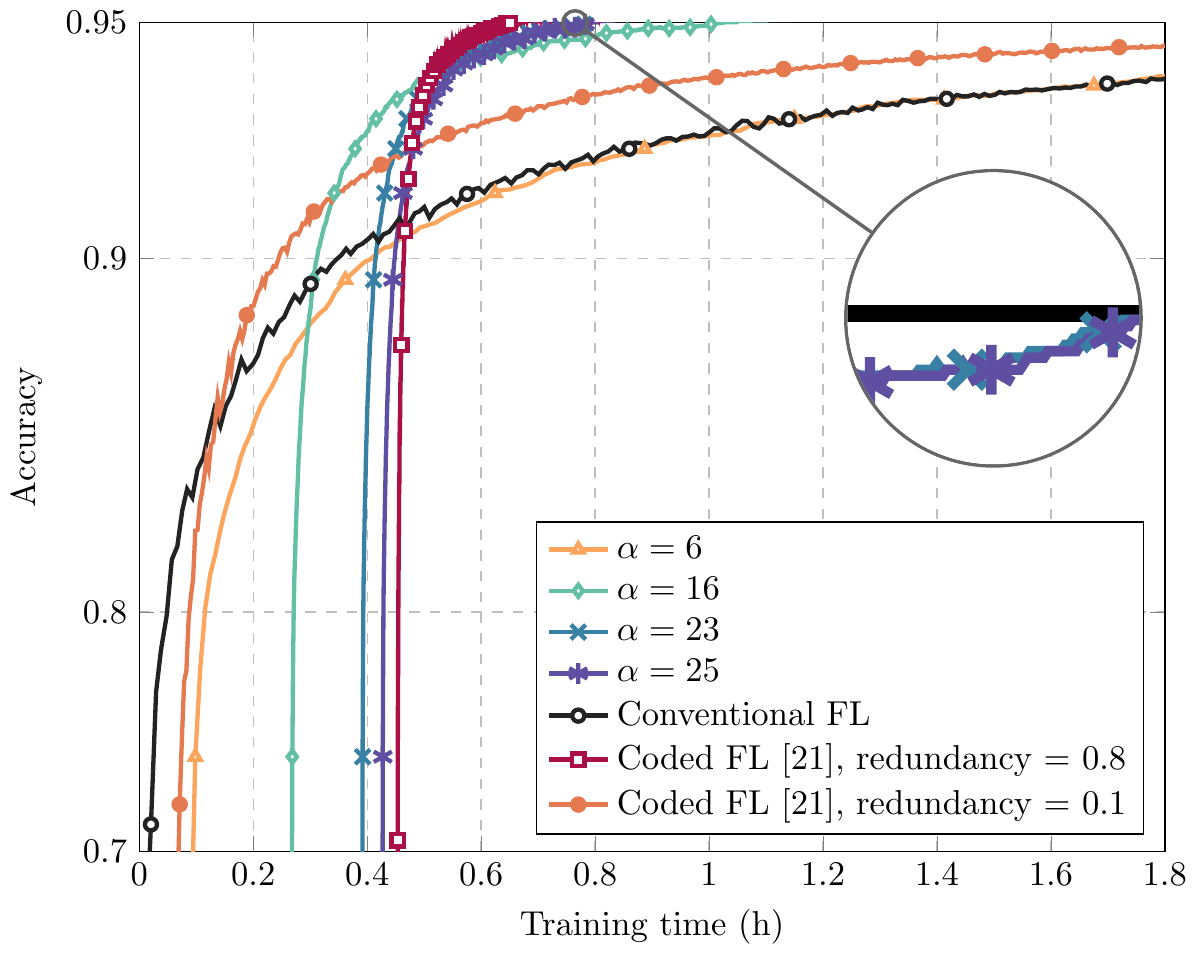}
        \caption{MNIST dataset}
        \label{Fig: TrainingTimeVSaccuracy_nu0.1}
    \end{subfigure}
    \begin{subfigure}[b]{0.45\textwidth}
        \centering
        \includegraphics[width=1\columnwidth]{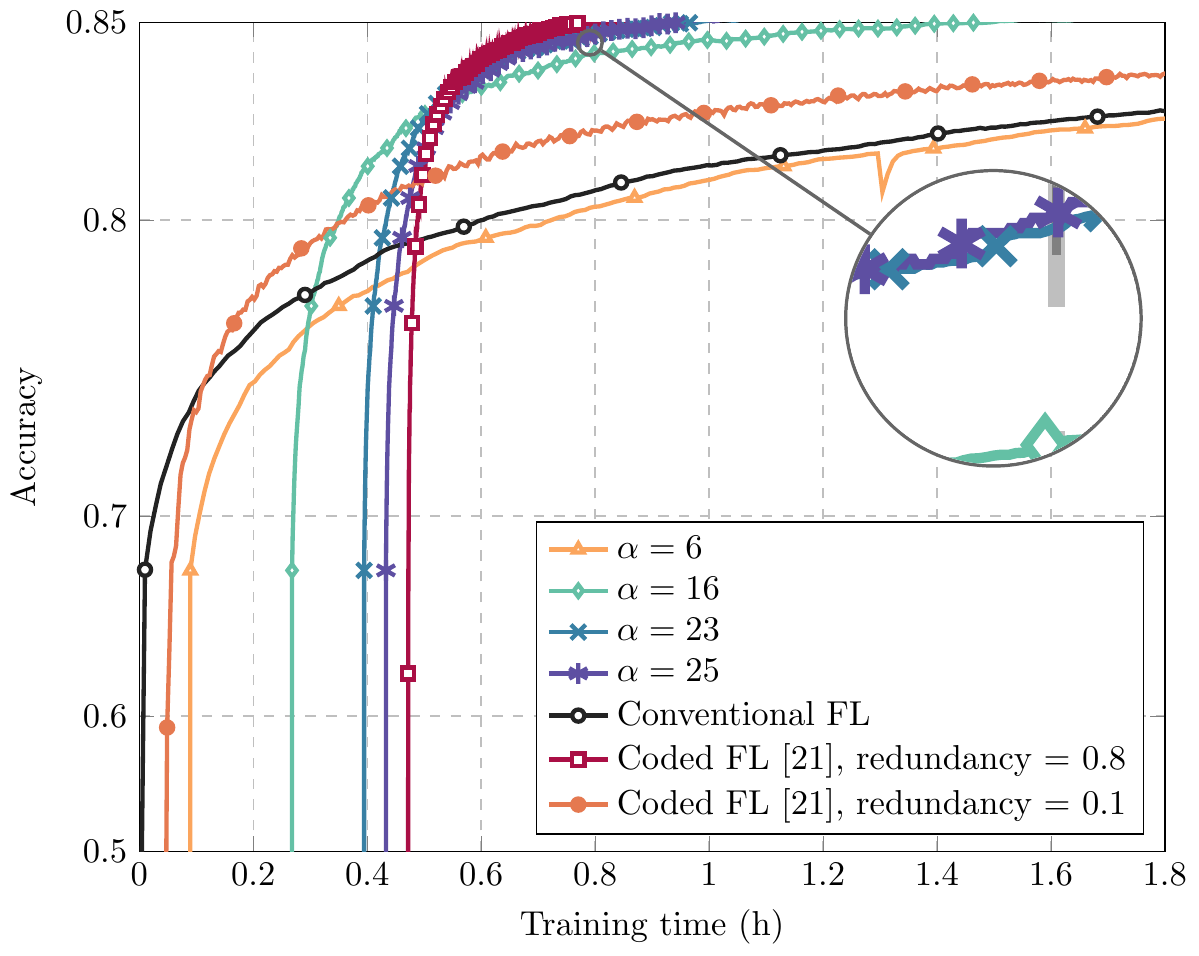}
        \vspace{-3ex}
        \caption{Fashion-MNIST dataset}
        \label{Fig: FMNIST_TrainingTimeVSaccuracy_nu0.1}
    \end{subfigure}
    \caption{Training time for the proposed coded FL scheme with different values of $\alpha$, the coded FL scheme in \cite{Pra21}, and conventional FL.}
   \vspace{-4ex}
\end{figure}

In \cref{Fig: TrainingTimeVSaccuracy_nu0.1}, we plot the training time of the proposed coded FL scheme and compare it with that of conventional FL and the scheme in \cite{Pra21} using the MNIST dataset.
For our scheme, we consider  $\alpha\in\{6, 16, 23, 25\}$. Note that $\alpha = 25$ corresponds to a replication scheme and all the padded data will be available at all devices. As \(\alpha\) increases, so does the time required to complete the encoding and sharing phase (note that there is no encoding and sharing phase for conventional FL). This induces a delay in the start of the training phase, which can be observed in the figure by the initial offset of the coded FL curves. However, once the sharing phase is completed, the time required to finish an epoch reduces as \(\alpha\) increases, as the central server only needs to wait for the gradients from the $D-\alpha+1$ fastest devices to perform the model update.
We see that the proposed coded FL with \(\alpha=23\) requires the least training time to achieve an accuracy of \(95\%\), yielding a speed-up of approximately \(6.6\) compared to conventional FL, where we have to wait for the slowest device in each epoch. For different target accuracy levels, different values of $\alpha$ will yield the lowest latency. If the target accuracy lies below $90\%$, it turns out that conventional FL outperforms the proposed scheme. Furthermore, too low values of $\alpha$, such as $\alpha = 6$, will never yield a lower latency for a given accuracy than conventional FL. The scheme in \cite{Pra21} trades off efficient training with privacy. 
To quantify the amount of parity data introduced, in \cite{Pra21}, the authors define a parameter $\delta$ as the amount of parity data per device over the total amount of raw data across devices.
Here, we choose two extreme values for $\delta$, namely $0.1$ and $0.8$. Note that the higher $\delta$ is, the more data is leaked to the central server. Our proposed  scheme  achieves a faster training time than the scheme in \cite{Pra21} with $\delta=0.1$ for an accuracy of $95\%$, while it achieves a slightly worse training time than the scheme in \cite{Pra21} with $\delta=0.8$. It is important to realize, though,  that a large $\delta$ goes against the spirit of FL: it leaks almost all data to the central server.

A similar behavior is observed  for the Fashion-MNIST dataset in \cref{Fig: FMNIST_TrainingTimeVSaccuracy_nu0.1}, for which  \(\alpha=25\) gives the best performance for an accuracy of \(85\%\) with a speed-up factor of approximately $9.2$ compared to conventional FL. However, if the target accuracy is between \(80\%\) and \(85\%\), nontrivial coding schemes (i.e., \(\alpha<25\)) perform best.

We still see gains when comparing our scheme to conventional FL where we drop the slowest devices at each epoch. However, due to space limitations, we do not include the results here.%

\section{Conclusion}
We proposed a novel coded FL scheme for linear regression that provides resiliency to straggling devices, while preserving the privacy level of conventional FL. The proposed scheme combines one-time padding\textemdash exploiting a fixed-point  arithmetic representation of the  data\textemdash to retain privacy and gradient codes to mitigate the effect of stragglers.
For a given target accuracy, the proposed scheme can be optimized to minimize the latency. For the MNIST dataset and an accuracy of $95\%$, our proposed coded FL scheme achieves a training speed-up factor of  \(6.6\) compared to conventional FL, while for the Fashion-MNIST dataset our scheme achieves a training speed-up factor of $9.2$ for an  accuracy of \(85\%\). Furthermore, our scheme yields comparable latency performance to the coded FL scheme in \cite{Pra21}, without incurring  the additional loss in privacy of this scheme. 
While the focus of this paper is in linear regression, the proposed scheme can also be applied to nonlinear optimization problems, e.g., classification, by transforming the dataset using kernel embedding.

\balance

\bibliographystyle{IEEEtran}
\bibliography{CFGD.bib}

\end{document}